\documentclass{article}

\usepackage[preprint]{nips_2018}

\usepackage{microtype}
\usepackage{graphicx}
\usepackage{subfigure}
\usepackage{booktabs} 

\usepackage[utf8]{inputenc} 
\usepackage[T1]{fontenc}    
\usepackage{hyperref}       
\usepackage{url}            
\usepackage{booktabs}       
\usepackage{amsfonts}       
\usepackage{nicefrac}       
\usepackage{microtype}      
\usepackage{amsthm}	
\usepackage{amsbsy} 
\usepackage{amsmath}
\usepackage{graphicx}
\usepackage{adjustbox}
\usepackage{bm}
\usepackage{wrapfig}
\usepackage{natbib}
\usepackage[textsize=tiny,textwidth=1.0cm]{todonotes}
\RequirePackage{algorithm}
\RequirePackage{algorithmic}

\usepackage{hyperref}

\newtheorem{lemma}{Proposition}
\newtheorem{sublemma}{Remark}[lemma]

\newcommand{\X}{\mathbf{X}}
\newcommand{\D}{\mathcal{D}}
\newcommand{\y}{\mathbf{y}}

\title{Combining model and parameter uncertainty in Bayesian Neural Networks}

\author{
  Aliaksandr Hubin \\
  SAMBA department\\
  Norwegian Computing Center\\
  Oslo, 0373 \\
  \texttt{aliaksandr.hubin@nr.no} \\
  \And
  Geir Storvik\\
  Department of Mathematics\\
  University of Oslo\\
  Oslo, 0316 \\
  \texttt{geirs@math.uio.no}
}
\hypersetup{draft}
\begin{document}
\maketitle
\begin{abstract}
Bayesian neural networks (BNNs) have recently regained a significant amount of attention in the deep learning community due to the development of scalable approximate Bayesian inference techniques. There are several advantages of using Bayesian approach: Parameter and prediction uncertainty become easily available, facilitating rigid statistical analysis. Furthermore, prior knowledge can be incorporated. However so far there have been no scalable techniques capable of combining both model (structural) and parameter uncertainty. In this paper we introduce the concept of model uncertainty in BNNs and hence make inference in the joint space of models and parameters. Moreover, we suggest an adaptation of a scalable variational inference approach with reparametrization of marginal inclusion probabilities to incorporate the model space constraints. Finally, we show that incorporating model uncertainty via Bayesian model averaging and Bayesian model selection allows to drastically sparsify the structure of BNNs.
\end{abstract}

\section{Introduction}

Bayesian neural networks represent a very flexible class of models, which are quite robust to overfitting. However they often remain heavily over-parametrized. There are several implicit approaches for BNNs sparsification through shrinkage of weights~\citep{blundell2015weight, molchanov2017variational,ghosh2018structured, neklyudov2017structured}. For example, \citet{blundell2015weight} suggest a mixture-of-Gaussians spike and slab prior on the weights and then perform fully factorisable mean-field variational approximation. \citet{ghosh2017model, louizos2017bayesian} independently generalize this approach by means of suggesting Horseshoe priors \citep{carvalho2009handling} for the weights, providing even stronger shrinkage and automatic specification of the mixture component variances required in~\citet{blundell2015weight}. \citet{molchanov2017variational} suggest that the interpretation of Gaussian dropout as performing variational approximations for a Bayesian neural network with log uniform prior over the weight parameters leads to improved sparsity in the latter. Some of these approaches show that the majority of the parameters can be pruned out from the network without significant loss of predictive accuracy. At the same time pruning is done in an implicit manner by deleting the weights via ad-hoc thresholding. 

In Bayesian model selection problems there have been numerous works showing efficiency and accuracy of model selection by means of introducing latent variables corresponding to different discrete model configurations, and then conditioning on their marginal posterior to both select the best sparse configuration and address the joint model-and-parameters-uncertainty explicitly \citep{george1993variable,Clyde:Ghosh:Littman:2010,Frommlet2012,hubin2018mode,Hubin2017,hubin2018deep}. For instance, \citet{hubin2018deep} address inference in the class of deep Bayesian regression models (DBRM), which generalizes the class of Bayesian neural networks. They show both good predictive performance of the obtained sparse models and the ability to recover meaningful complex nonlinearities. 
The approach suggested in \citet{hubin2018deep} is based on adaptations of Markov chain Monte Carlo (MCMC) and does not scale well to large highdimensional data samples. \citet{louizos2017bayesian} also warn about the complexity of explicit discretization of model configuration within BNNs, as it causes an exponential explosion with respect to the total number of parameters, and hence infeasibility of inference for highdimensional problems. At the same time, \citet{polson2018posterior} encourage the use of spike and slab approach in BNNs from a theoretical standpoint.

\citet{logsdon2010variational, carbonetto2012scalable} suggested a fully-factorized variational distribution capable of efficiently and precisely "linearizing" the problem of Bayesian model selection in the context of \emph{linear} models with an ultrahigh number of potential covariates, typical for genome wide association studies (GWAS). In the discussion to his PhD thesis \citet{hubin2018thesis} proposed combining the approaches of \citet{logsdon2010variational, carbonetto2012scalable} and  \citet{graves2011practical}  for scalable approximate Bayesian inference on the joint space of models and parameters in deep Bayesian regression models. We develop this idea further in this article. 

More specifically, we introduce a formal Bayesian approach for jointly taking into account \textit{model} (structural) \textit{uncertainty} and \textit{parameter uncertainty} in BNNs. The approach is based on introducing latent binary variables corresponding to inclusion-exclusion of particular weights within a given architecture. Using Bayesian formalization in the space of models allows to adapt the whole machinery of Bayesian inference in the joint model-parameter settings, including \textit{Bayesian model averaging} (BMA) (across all models) or \textit{Bayesian model selection} (BMS) of one \textit{“best”} model with respect to some model selection criterion \citep{claeskens2008model}. In this paper we study BMA as well as the \textit{median probability} model \citep{barbieri2004optimal, barbieri2018median} and \textit{posterior mean} model based inference for BNNs. Spasifying properties of BMA and the median probability model are also addressed. Finally, following \citet{hubin2018thesis} we will link the obtained \textit{marginal inclusion probabilities} to \textit{binary dropout} rates, which gives proper probabilistic reasoning for the latter. The suggested inference approach is based on scalable stochastic variational inference. 

The approach has similarities to binary dropout that has become very popular~\citep{srivastava2014dropout}. However, while standard binary dropout can only be seen as a Bayesian approximation to a Gaussian process model where only parameter estimation is taken into account~\citep{Gal2016Uncertainty}, our approach also explicitly models structural uncertainty. In this sense it is closely related to Concrete dropout \citep{gal2017concrete}. However, the model proposed by \citet{gal2017concrete} does not allow for BMS: The median probability model will either select all weights or nothing due to a strong assumption on having the same dropout probabilities for the whole layer. Furthermore, it uses variational approximations, which were not studied in the model uncertainty context. 

The rest of the paper is organized as follows: The class of BNNs and the corresponding model space are  mathematically defined in Section~\ref{section2}. In Section~\ref{section3} we describe the algorithm for making inference on the suggested class of models using reparametrization of marginal inclusion probabilities. In Section~\ref{section4} the suggested approach is applied to a classical benchmark data set MNIST and also to a data set FMNIST, there we also compare the results with some of the existing approaches for inference on BNNs. Finally, in Section~\ref{section5} some conclusions and suggestions for further research are given. Additional results, discussions, and proofs can be found in the web supplement to the paper.


\section{The model}\label{section2}
A neural network model links observations $\y\in \mathcal{R}^r$ and explanatory variables $\X$  via a probabilistic functional mapping of the form~\eqref{eq:ann}, with a possibly multidimensional mean parameter $\boldsymbol \mu_i = \boldsymbol \mu_i(\boldsymbol x_i), \boldsymbol \mu_i \in \mathcal{R}^r, i \in \{1,...,n\}$, which can be written as: 
\begin{align}
\boldsymbol y_i \sim \mathfrak{f}\left(\boldsymbol \mu_i, \phi \right),\label{eq:ann} 
\end{align}
where $\phi$ is a dispersion parameter. To construct the vector of mean parameters $\boldsymbol \mu_i$ of the  distribution of interest, one builds a sequence of building blocks of hidden layers through semi-affine transformations:
\begin{align}
z^{(l+1)}_{ij} =&\sigma^{(l)}_j\left(\gamma^{(l)}_{0j}\beta^{(l)}_{0j} + \sum_{k=1}^{p^{(l)}}\gamma^{(l)}_{kj}\beta^{(l)}_{kj}z^{(l)}_{ik}\right), l=1,...,L, j=1,...,p^{(l+1)}.\label{eq:neuron}
\end{align}
with $\mu_{ij}=z_{ij}^{(L)}$. Here $L$ is the number of layers, $p^{(l)}$ is the number of nodes within the layer while $\sigma^{(l)}_{j}$ are univariate functions (further referred to as \textit{activation functions}).
Further, $\beta^{(l)}_{kj}\in\mathcal{R}$ are the weights (slope coefficients) for the inputs $z^{(l)}_{ik}$ of the $l$-th layer (note that $z^{(1)}_{ik} = x_{ik}$; for $k=0$ we obtain the intercepts or the bias terms). Finally, $\gamma^{(l)}_{kj}\in \{0,1\}$ are latent binary indicators switching the corresponding weights on and off.

In our notation we explicitly differentiate between discrete model configurations defined by the vectors 
$\boldsymbol\gamma =\{\gamma_{kj}^{(l)},j=1,..,p^{(l+1)},k=0,...,p^{(l)},l=1,...,L\}$ 
(further referred to as models) constituting the model space $\Gamma$ and parameters of the models, conditional on these configurations $\boldsymbol \theta|\boldsymbol\gamma = \{\boldsymbol\beta,\phi|\boldsymbol\gamma\}$, where only those $\beta_{kj}^{(l)}$ for which $\gamma_{kj}^{(l)}=1$ are included. 
This approach is a rather standard (in statistical science literature) way to explicitly specify the model uncertainty in a given class of models and is used in \citet{Clyde:Ghosh:Littman:2010,Frommlet2012,hubin2018mode,Hubin2017,hubin2018deep}. 
A Bayesian approach is obtained by specification of model priors $p(\boldsymbol\gamma)$ and parameter priors for each model $p(\boldsymbol\beta|\boldsymbol\gamma)$. If the dispersion parameter is present in the distribution of the outcomes, one also has to define
$p(\phi|\boldsymbol\gamma) = d_{\phi}(\phi)$. Many kinds of priors on $p(\boldsymbol\beta|\boldsymbol\gamma)$ can be considered, including the mixture of Gaussians prior \citep{blundell2015weight}, Horseshoe prior \citep{ghosh2017model,louizos2017bayesian}, or mixtures of g-priors \citep{li2018mixtures}, but for simplicity of representation we consider the multivariate Gaussian with the diagonal covariance matrix weight prior combined with independent Bernoulli priors for the latent inclusion indicators \citep{Clyde:Ghosh:Littman:2010}:
\begin{equation}
p(\beta^{(l)}_{kj}|\gamma^{(l)}_{kj}=1) = N(0,\sigma_{\beta}^2),\quad p(\gamma^{(l)}_{kj}=1)=\text{Bernoulli}(\psi).\label{eq:prior.beta}
\end{equation}
The parameter  $\psi\in(0,1)$ is the penalty for including the weight $\beta^{(l)}_{kj}$ into the model. To induce AIC-type model complexity penalization, one typically uses $\psi = \exp{(-2)}$, whilst for BIC-type penalization one uses  $\psi =\exp{(-2 \log n)}$, where $n$ is the sample size.

\section{Bayesian inference}\label{section3}

The main goal of inference with uncertainty in both models and parameters is to infer the posterior marginal distribution of some parameter of interest $\Delta$ 
(for example probability of a new observation $y^*$ conditional on new covariates $\boldsymbol{x}^*$) based on data $\D$:
\begin{align}\label{eq:pred.delta}
p(\Delta|\D) = \textstyle\sum_{\boldsymbol\gamma \in \Gamma}\int_{\Theta}p(\Delta|\boldsymbol\theta,\boldsymbol\gamma,\D)p(\boldsymbol\theta,\boldsymbol\gamma|\D)\text{d}\boldsymbol\theta.
\end{align}
The posterior joint distribution of models and parameters $p(\boldsymbol\theta,\boldsymbol\gamma|\D)$ will be approximated by means of scalable variational inference~\citep{graves2011practical} 
for the joint parameter-model setting. 
  Consider the following variational families, parameterized with $\boldsymbol\eta$, with independence across weight components:
\begin{align}
q_{{\boldsymbol\eta}}(\boldsymbol\theta,\boldsymbol\gamma) = q_{\boldsymbol\eta_0}(\phi) \textstyle\prod_{l=1}^{L-1}\textstyle\prod_{j=1}^{p^{(l+1)}}\textstyle\prod_{k=0}^{p^{(l)}}q_{{\boldsymbol\eta^{(l)}_{kj}}}(\beta^{(l)}_{kj},\gamma^{(l)}_{kj}),\label{eq:varfactms}
\end{align}
where the factorized components $q_{{\boldsymbol\eta^{(l)}_{kj}}}(\beta^{(l)}_{kj},\gamma^{(l)}_{kj})$ follow the form suggested (in a simpler setting) by \citet{logsdon2010variational, carbonetto2012scalable}:
\begin{align}
q_{{\boldsymbol\eta^{(l)}_{kj} }}\left(\beta^{(l)}_{kj},\gamma^{(l)}_{kj}\right) =\begin{cases} \alpha^{(l)}_{kj} N\left(\mu^{(l)}_{kj},{\sigma^2}^{(l)}_{kj}\right), \text{ if $\gamma^{(l)}_{kj}=1$;}\\
\left(1-\alpha^{(l)}_{kj}\right)\delta_0(\beta^{(l)}_{kj}), \text{ if $\gamma^{(l)}_{kj}=0$.}
\end{cases}
\end{align}
Here $\delta_0(\cdot)$ is the delta mass or "spike" at zero and $\boldsymbol\eta^{(l)}_{kj} = (\mu^{(l)}_{kj},{\sigma^2}^{(l)}_{kj},\alpha^{(l)}_{kj})$. Thus, with probability $\alpha^{(l)}_{kj}\in \mathcal{R}_{[0,1]}$, the posterior of parameters of a weight $\beta^{(l)}_{kj}$ will be approximated by a normal distribution with some mean and variance ("slab"), and otherwise the edge is considered to have no effect on the observations $\y$. Hence $\alpha^{(l)}_{kj}$ will approximate the marginal inclusion probability of the weight $\beta^{(l)}_{kj}$.
Note that while the \emph{variational} distribution $q_{\bm\eta}$ is similar to the dropout approach, the \emph{target} distribution, which we aim at approximating, is different in the sense of including the binary variables $\{\gamma^{(l)}_{kj}\}$ as well. Hence our marginal inclusion probabilities can serve as a particular case of \textit{dropout} rates with a proper probabilistic interpretation in terms of structural model uncertainty.

To perform variational approximation of the posterior we aim at minimizing the Kullback-Leibler divergence between the variational family distribution and the posterior distribution.
\begin{align}
\text{KL}(q_{\boldsymbol\eta}(\boldsymbol\theta,\boldsymbol\gamma)&||p(\boldsymbol\theta,\boldsymbol\gamma|\D))
 = \sum_{\boldsymbol\gamma \in \Gamma}\int_{\Theta}q_{\boldsymbol\eta}(\boldsymbol\theta,\boldsymbol\gamma)\log \tfrac{q_{\boldsymbol\eta}(\boldsymbol\theta,\boldsymbol\gamma)}{p(\boldsymbol\theta,\boldsymbol\gamma|\D)}\text{d}\boldsymbol\theta,\label{eq:KL}
\end{align}
 with respect to the variational parameters $\boldsymbol\eta$. Constraints on $\alpha^{(l)}_{kj}$ are incorporated by means of the reparametrization $
\alpha^{(l)}_{kj} = ({1+\exp(-\omega^{(l)}_{kj})})^{-1}, \text{ where } \omega^{(l)}_{kj}\in\mathcal{R}.$
For numerical stability we also reparameterize the standard deviations of the weights $
\sigma^{(l)}_{kj} =\log(1+\exp(\rho^{(l)}_{kj}))$. In Proposition~\ref{lone} we show minimization of the divergence~\eqref{eq:KL} to be equivalent to maximization of the evidence lower bound $\mathcal{L}_{VI}(\boldsymbol\eta)$.
\begin{lemma}\label{lone}
Minimization of $\text{KL}\left(q_{\boldsymbol\eta}(\boldsymbol\theta,\boldsymbol\gamma)||p(\boldsymbol\theta,\boldsymbol\gamma|\D)\right)$ is equivalent to maximization of
the \textit{evidence  lower bound} (ELBO)
\begin{align*}
\mathcal{L}_{VI}(\boldsymbol\eta):=& \textstyle\sum_{\boldsymbol\gamma \in \Gamma}\textstyle\int_{\Theta}q_{\boldsymbol\eta}(\boldsymbol\theta,\boldsymbol\gamma)\log p(\D|\boldsymbol\theta,\boldsymbol\gamma) \text{d}\boldsymbol\theta-\text{KL}\left(q_{\boldsymbol\eta}(\boldsymbol\theta,\boldsymbol\gamma)||p(\boldsymbol\theta,\boldsymbol\gamma)\right) 
\end{align*}
\textbf{Proof} of Proposition~\ref{lone} is given in Section A of the  web supplement. 
\end{lemma}

Following \citet{blundell2015weight} we approximate the objective $\mathcal{L}_{VI}$ with an unbiased estimate based on mini-batching. Note that the observations are assumed conditionally independent, which means $\sum_{\boldsymbol\gamma \in \Gamma}\int_{\Theta}q_{\boldsymbol\eta}(\boldsymbol\theta,\boldsymbol\gamma)\log{p(\D|\boldsymbol\theta,\boldsymbol\gamma)}\text{d}\boldsymbol\theta=\sum_{i = 1}^{n}\sum_{\boldsymbol\gamma \in \Gamma}\int_{\Theta}q_{\boldsymbol\eta}(\boldsymbol\theta,\boldsymbol\gamma)\log{p(\boldsymbol{y_i}|\boldsymbol{x_i},\boldsymbol\theta,\boldsymbol\gamma)}\text{d}\boldsymbol\theta$. We will hence sample mini-batches $S$ of size $N$ from the full data, giving the unbiased estimate
\begin{align*}
\hat{\mathcal{L}}_{VI}(\boldsymbol\eta) =& \tfrac{n}{N}\textstyle\sum_{i\in S}\textstyle\sum_{\boldsymbol\gamma \in \Gamma}\textstyle\int_{\Theta}q_{\boldsymbol\eta}(\boldsymbol\theta,\boldsymbol\gamma)\log{p(\boldsymbol{y_i}|\boldsymbol{x_i},\boldsymbol\theta,\boldsymbol\gamma)}\text{d}\boldsymbol\theta-{\text{KL}\left({q_{\boldsymbol\eta} (\boldsymbol\theta,\boldsymbol\gamma)}||{p(\boldsymbol\theta|\boldsymbol\gamma)p(\boldsymbol\gamma)}\right)}.
\end{align*}
However, the cardinality of $\Gamma$ is $2^q$, where $q$ is the total number of potential weights in the neural network, hence iterating through all of the models in $\Gamma$ for every iteration of stochastic gradient descent optimization is infeasible for reasonably complex Bayesian neural networks. It is also infeasible to integrate over all of the weights for a given configuration. To resolve this, we adopt a yet another unbiased Monte-Carlo based approximation:
\begin{align}
\widetilde{{\mathcal{L}}}_{VI}(\boldsymbol\eta) =&\tfrac{1}{M}\left[\textstyle\sum_{m=1}^M\frac{n}{N}\textstyle\sum_{i\in S}\textstyle\log{p(\boldsymbol{y_i}|\boldsymbol{x_i},\boldsymbol\theta^{(m)},\boldsymbol\gamma^{(m)})}-
{\log \tfrac{q_{\boldsymbol\eta} (\boldsymbol\theta^{(m)},\boldsymbol\gamma^{(m)})}{p(\boldsymbol\theta^{(m)}|\boldsymbol\gamma^{(m)})p(\boldsymbol\gamma^{(m)})}}\right],\label{estloss}
\end{align}
where $\left(\boldsymbol\theta^{(m)},\boldsymbol\gamma^{(m)}\right) \sim q_{\boldsymbol\eta}(\boldsymbol\theta,\boldsymbol\gamma)$ for $m=1,...,M$. 

Stochastic gradient based methods for optimization require calculation of an unbiased estimate of the gradient \citep{bottou2018optimization}. One can further use sampling to obtain such an estimate \citep{graves2011practical,blundell2015weight}. Some extra care is needed in this case, due to the discrete $\bm\gamma$ variables. In Proposition~\ref{ltwo} we suggest a generalization of~\citet{blundell2015weight} and \citet{mnih2014neural}, allowing to make inference in the joint space of models and parameters. Remark~\ref{r21} allows to use control variates \citep{weaver2001optimal} to reduce the variance of $\widetilde{\nabla} {\mathcal{L}}_{VI}(\boldsymbol\eta)$ \citep{mnih2014neural}. 
\begin{lemma}\label{ltwo}
Assume $\left(\boldsymbol\theta^{(m)},\boldsymbol\gamma^{(m)}\right) \sim q_{\boldsymbol\eta}(\boldsymbol\theta,\boldsymbol\gamma)$ for  $m \in \{1,...,M\}$ and
$S$ is a random subset of $\{1,...,n\}$ of size N. Then an
unbiased estimator for the gradient of ${\mathcal{L}}_{VI}(\boldsymbol\eta)$  is given by:
\begin{align}
\widetilde{\nabla} {\mathcal{L}}_{VI}(\boldsymbol\eta)
=& \frac{1}{M}\textstyle\sum_{m=1}^M\frac{n}{N}\textstyle\sum_{i\in S}\nabla\log{p(\boldsymbol{y_i}|\boldsymbol{x_i},\boldsymbol\theta,\boldsymbol\gamma)}-\frac{1}{M}\sum_{m=1}^M\nabla \log \tfrac{q_{\boldsymbol\eta} (\boldsymbol\theta,\boldsymbol\gamma)}{p(\boldsymbol\theta,\boldsymbol\gamma)},\label{eq:grad.LVI}
\end{align}
\end{lemma}
\begin{sublemma}\label{r21}

Without inducing bias, a term $l_{\boldsymbol{\eta}}(\boldsymbol\psi,\boldsymbol{x}):=\sum_{m=1}^{M}\nabla \log q_{\boldsymbol\eta}(\boldsymbol\theta^{(m)},\boldsymbol\gamma^{(m)})C_{\boldsymbol\psi}(\boldsymbol{x})$ can be added to $\widetilde\nabla {{\mathcal{L}}}_{VI}(\boldsymbol\eta)$.

\textbf{Proofs} of Proposition~\ref{ltwo} and Remark~\ref{r21} are given in Section A of the web supplement. 
\end{sublemma}

Algorithm~\ref{alg:dsvi}  describes one iteration of a \textit{doubly stochastic} variational inference approach where updating is performed on the parameters $\mu_{kj}^{(l)},\rho_{kj}^{(l)},\omega_{kj}^{(l)}$ in the transformed space. 
The set $\mathcal{B}$ is the collection of all combinations $j,k,l$  in the network. Note that in the suggested algorithm partial derivatives with respect to marginal inclusion probabilities, as well as mean and standard deviation terms of the weights are shared and coincide with the gradients found by the usual backpropagation algorithm on a neural network. This algorithm assumes no dispersion parameter $\phi$, but can be easily generalized to include it.
Once the estimates of the parameters $\widehat{\boldsymbol\eta}$ of the variational approximating distribution are obtained, there are several ways to proceed with the inference. Algorithm~\ref{alg:pred} describes a procedure for inference on $p(\Delta|\D)$ based on the marginal posterior predictive distribution~\eqref{eq:pred.delta}, 
taking uncertainty in both the model structure $\bm\gamma$ and the parameters $\bm\beta$ into account.  \begin{minipage}[!t]{1\linewidth}\centering
\begin{minipage}[!t]{0.53\linewidth}
\begin{algorithm}[H]
\footnotesize
   \caption{\footnotesize Doubly stochastic variational inference step}
   \label{alg:dsvi}
   \tiny
\begin{algorithmic}
   \STATE \textbf{sample} $N$ indices uniformly from $\{1,...,n\}$ defining $S$;
   \FOR{$m$ {\bfseries in} $\{1,...,M\}$}
   \FOR {$(k,j,l)\in\mathcal{B}$}
   \STATE \textbf{set} $\alpha^{(l)}_{kj} =  \left(1+\exp({-\omega^{(l)}_{kj}})\right)^{-1}$, $\sigma^{(l)}_{kj} = \log\left(1+\exp({\rho^{(l)}_{kj}})\right)$;
   \STATE \textbf{sample} $\gamma^{(l)}_{kj} \sim \text{Bernoulli}(\alpha^{(l)}_{kj})$;
   \STATE \textbf{sample} $\beta^{(l)}_{kj} \sim (\mu^{(l)}_{kj} +{\sigma}^{(l)}_{kj}N(0,1))\text{I}(\gamma^{(l)}_{kj}=1)$;
   \ENDFOR
   \ENDFOR
    \STATE \textbf{set} ${\boldsymbol\eta} \leftarrow {\boldsymbol\eta} + a\widetilde{\nabla}{{\mathcal{L}}}_{VI}(\boldsymbol\eta)$ with  $\widetilde{\nabla}{{\mathcal{L}}}_{VI}(\boldsymbol\eta)$ from~\eqref{eq:grad.LVI};
\end{algorithmic}
\end{algorithm}
\end{minipage}
\begin{minipage}[!t]{0.45\linewidth}
\begin{algorithm}[H]
\footnotesize
\caption{\footnotesize Prediction by model averaging}
   \label{alg:pred}
   \tiny
\begin{algorithmic}
   \STATE \textbf{foreach} prediction $\Delta$ of interest \textbf{do}
   \FOR{$r$ {\bfseries in} $1,...,R$}
    \FOR {$(k,j,l)\in\mathcal{B}$}
   \STATE \textbf{sample} ${\gamma^{(l)}_{kj}} \sim \text{Bernoulli}(\alpha^{(l)}_{kj})$;
   \STATE \textbf{sample} $\beta^{(l)}_{kj} \sim N(\mu^{(l)}_{kj},{\sigma^2}^{(l)}_{kj})\text{I}(\gamma^{(l)}_{kj}=1)$;
   \ENDFOR
   \STATE \textbf{calculate} $p^{(r)}(\Delta)=p(\Delta|\boldsymbol\beta,\boldsymbol\gamma,\D)$;
    \ENDFOR
   \STATE \textbf{set} $\hat{p}(\Delta|\D) = \frac{1}{R}\sum_{r=1}^{R}p^{(r)}(\Delta)$.
   \STATE \textbf{end foreach}
\end{algorithmic}
\end{algorithm}
\end{minipage}
\end{minipage}

A bottleneck of Algorithm~\ref{alg:pred} is that we have to both sample from a huge approximate posterior distribution of parameters and models and keep all of the components of $\widehat{\boldsymbol\eta}$ stored during the inference, which might be computationally inefficient. The \textit{posterior mean} based model \citep{wasserman2000bayesian}, which sets $\beta^{(l)}_{kj} \leftarrow \hat E\{\beta^{(l)}_{kj}|\D\}$ and $\alpha^{(l)}_{kj} \leftarrow 1$ or the \textit{median probability} model \citep{barbieri2004optimal}, which sets $\alpha^{(l)}_{kj} \leftarrow \text{I}(\alpha^{(l)}_{kj}>0.5)$, can be used as simplifying alternatives. Both approaches specify one model $\hat{\bm\gamma}$ but the former model results in a dense model while the latter results in a sparse one. 

Given a specific model $\hat{\bm\gamma}$ there are several options with respect to $\bm\beta$: Simulation conditional on $\hat{\bm\gamma}$ as in Algorithm~\ref{alg:pred} or inserting the posterior means conditional on $\hat{\bm\gamma}$. A main benefit with the latter is that a single set $(\hat{\bm\gamma},\hat{\bm\beta})$ is predefined so that calculation of only one term $p(\Delta|\hat{\boldsymbol\beta},\hat{\boldsymbol\gamma},\D)$ needs to be done in the prediction step. In both of these cases, one can either use directly the learned variational approximation or perform post-training of the parameters. By post-training we mean that only distributions of the parameters of the models (means and variances of the weights and dispersion parameter of the distribution) are optimized, whilst the distributions of the model configurations are fixed. This might improve inferential properties of the posterior predictive distribution and reduce its variance. We provide a comparison of several alternatives in Section~\ref{section4} and give a further discussion of the simplifying inference techniques in Section~\ref{OtherInf} of the web supplement.

\section{Applications}\label{section4}

In this section we will address the classification of MNIST \citep{lecun1998mnist} images and classification of fashion-MNIST \citep[FMNIST][]{xiao2017fashion} images. Both of the datasets comprise of 28x28 grayscale images of $70\,000$ items from 10 categories, with $7\,000$ images per category. The training sets consist of $60\,000$ images and the test sets have $10\,000$ images. 

\paragraph{Experimental design} For both of the data sets we addressed a dense neural network with the ReLU activation function, multinomially distributed observations with 10 classes and 784 input explanatory variables (pixels). The network has 3 hidden layers with 400, 600 and 600 neurons correspondingly. Priors for the parameters and model indicators were chosen according to~\eqref{eq:prior.beta} with
$\psi=\exp(-2)$ and $\sigma^2_\beta = 1$.
 The inference was performed by means of the suggested doubly stochastic variational inference approach on 250 epochs with batch size 100. We used the ADAM stochastic gradient descent optimization \citep{kingma2014adam} with $a_\mu = a_\rho = 0.0001, a_\omega = 0.1$ (allowing the $a$ parameter in the last step of Algorithm~\ref{alg:dsvi} to have separate values for the different sets of parameters). When post-training the parameters, either with fixed marginal inclusion probabilities or with the median probability model, we ran additional $50$ epochs of the optimization routine 
 with $a_\mu = a_\rho = 0.0001$, and $a_\omega = 0$. 
 In  Algorithm~\ref{alg:pred} we used both $R=1$ and $R=10$.

\newcounter{num}
\setcounter{num}{10}
\ifodd\value{num} 
\begin{minipage}[b]{1\linewidth}\centering
\begin{minipage}[b]{0.7\linewidth}
\vspace{0.1cm}
\begin{table}[H]
 \caption{\small Inference possibilities. \textbf{SM} is a single sample, \textbf{MA} - model (sample) averaging, \textbf{MN} - posterior mean based inference, \textbf{MED} - selecting the median probability model, \textbf{WAIC}, \textbf{DIC}, \textbf{FIC} - selecting w.r.t. the corresponding criterion, \textbf{ADHOC} - add hoc model selection, \textbf{PT} - post-training of the parameter, \textbf{PE} - point estimates of the predictions, \textbf{CI} - credible intervals for the predictions.}
  \label{t0}
  \tiny
  \centering
{
  \begin{tabular}{l@{\hspace{1.0cm}}c@{\hspace{0.7cm}}c@{\hspace{0.7cm}}c@{\hspace{0.7cm}}c@{\hspace{0.7cm}}c@{\hspace{0.7cm}}c}
   \toprule
Method&Full BNN&Gauss.&Mixt.&Concr.&Hors.&Studied\\\hline
&&\multicolumn{4}{l}{\hspace{0mm} \textbf{Dense}}\\\hline
$\text{SM}$&Joint&Par&Par&Joint&Par&\textbf{Yes}\\
$\text{MA}$&Joint&Par&Par&Joint&Par&\textbf{Yes}\\
$\text{MN}$&Joint&Par&Par&Joint&Par&\textbf{Yes}\\
\hline
&&\multicolumn{4}{l}{\hspace{0mm} \textbf{Model selection}}\\\hline
$\text{MED}$&+&-&-&-&-&\textbf{Yes}\\
$\text{WAIC, DIC, FIC}$&+&-&-&-&-&\textbf{No}\\
$\text{PRUNE}$&?&?&?&?&+&\textbf{Yes}\\
\hline
&&\multicolumn{4}{l}{\hspace{0mm} \textbf{Post training}}\\\hline
$\text{PT}$&MA/MS&-&-&MA&MS&\textbf{Yes}\\
\hline
&&\multicolumn{4}{l}{\hspace{0mm} \textbf{Inference}}\\\hline
$\text{PE}$ (Acc. All)&Joint&Par&Par&Joint&Par&\textbf{Yes}\\
$\text{CI}$ (Acc. 95\%)&Joint&Par&Par&Joint&Par&\textbf{Yes}\\
\bottomrule
  \end{tabular}
  }
\end{table}
\end{minipage}
\hspace{0.5cm}
\begin{minipage}[b]{0.25\linewidth}
\begin{table}[H]
 \caption{\small Medians and standard deviations of the average (per layer) marginal inclusion probability (see the text for definition) for our model for both MNIST and FMNIST data across 10 simulations.}
 \label{t3}
  \tiny
  \centering
  \begin{tabular}{l@{\hspace{0.4cm}}c@{\hspace{0.4cm}}c}
   \toprule
\multicolumn{3}{l}{\textbf{MNIST data}}\\
\hline
Layer& Med.  & SD.\\\hline
$\rho(\gamma^{(1)}|\D)$&0.0520&0.0005\\
$\rho(\gamma^{(2)}|\D)$&0.0598&0.0003\\
$\rho(\gamma^{(3)}|\D)$&0.2217&0.0064\\
\hline
\multicolumn{3}{l}{\textbf{FMNIST data}}\\
\hline
Layer& Med.  & SD.\\\hline
$\rho(\gamma^{(1)}|\D)$&0.0665&0.0004\\
$\rho(\gamma^{(1)}|\D)$&0.0613&0.0005\\
$\rho(\gamma^{(1)}|\D)$&0.2013&0.0051\\
\bottomrule
  \end{tabular}
\end{table}
\end{minipage}
\vspace{0.1cm}
 \end{minipage}
 \fi
We then evaluated accuracies (\textbf{Acc} - the proportion of the correctly classified images) for each vector of predictions and recorded their median. 
Accuracies based on a single sample ($R=1$) from the median probability model and the model obtained by inserting the posterior mean of the parameters were also obtained. Finally, accuracies based on post-training of the parameters with fixed marginal inclusion probabilities and post-training of the median probability model were evaluated. For the cases when model averaging is addressed ($R=10$) we are additionally reporting accuracies when classification is only performed when the maximum model averaged class probability exceeds 95\%  as suggested by \citet{posch2019variational}. Otherwise, a doubt decision is made~\citep[][sec 2.1]{ripley2007pattern}. In this case we both
report the accuracy within the classified images  as well as the number of classified images.
Finally, in column \textbf{Density} of Table~\ref{t1} we are reporting the median of the overall density level (proportion of used at least once weights in the prediction stage to the total number of weights in a BNN), for different approaches. 
To guarantee reproducibility, summaries across 10 independent runs of the described experiment $s \in \{1,...,10\}$ were computed for all of these statistics. Summaries of these statistics are reported in Table~\ref{t1}.
Estimates of the marginal inclusion probabilities~$\hat{p}(\gamma_{kj}^{(l)}=1|\D)$ based on the suggested variational approximations were also computed for all of the weights. In order to compress the presentation of the results we only present the mean marginal inclusion probabilities for all of the layers $l$ as $\rho(\gamma^{(l)}|\D) := \frac{1}{p^{(l+1)}p^{(l)}}\sum_{kj}\hat{p}(\gamma_{kj}^{(l)}=1|\D)$, summarized in Table~\ref{t3}.

In addition to \textit{our approach} (denoted as \textbf{Full BNN, Gaussian priors} in Table~\ref{t1}) we also used several \textit{baselines}. In particular, we addressed a standard \textbf{Dense BNN with Gaussian priors} \citep{graves2011practical}, which is a particular case of our original model with all $\gamma^{(l)}_{kj}$ being fixed and equal to 1. This model is important in measuring how much of the predictive power we might loose due to introducing sparsity.
Furthermore, we report the results for a \textbf{Dense BNN with mixture priors} with two Gaussian components of the mixtures \citep{blundell2015weight} with probabilities 0.5 for each and variances equal to 1 and $e^{-6}$ correspondingly. Additionally we have addressed two popular sparsity inducing approaches, in particular, \textbf{BNN with Concrete dropout} \citep{gal2017concrete} and \textbf{BNN with Horseshoe priors} \citep{louizos2017bayesian}. All of the baseline methods have, similarly to the full BNN, 3 hidden layers with 400, 600 and 600 neurons correspondingly. They were trained for 250 epochs with Adam optimizer ($a=0.0001$) and batch size equal to 100. 
For the BNN with Horseshoe priors we are reporting statistics separately before and after ad-hoc pruning of the weights. Post-training (when necessary) was performed for additional 50 epochs. 

\begin{figure}[t]
\centering
\includegraphics[width=0.49\linewidth]{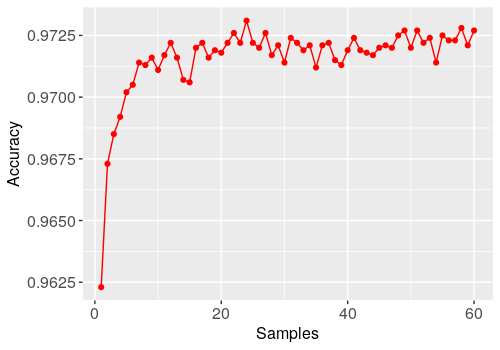}
\includegraphics[width=0.49\linewidth]{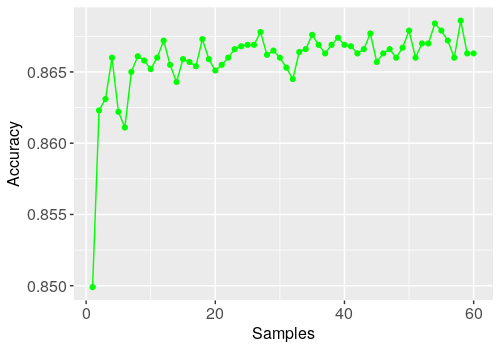}
\caption{Accuracy of predictions w.r.t. the number of samples from the joint posterior of models and parameters $R$ on the MNIST data (left) and FMNIST data (right) for experiment run $s=10$.}\label{Fig:predsamples}
\end{figure}

\begin{table}[t]
 \caption{\small Performance metrics  for the MNIST data (left) and FMNIST data (right) for the compared different Bayesian approaches to BNN. For $\bm\gamma$, SIM corresponds to that the $\gamma$'s are sampled from the posterior, ALL that all $\gamma$'s are put to 1, MED - that the median probability model is selected, PRN indicates add-hoc weight pruning.
 For $\bm\beta$, SIM corresponds to that the $\beta$'s are simulated from the posterior, EXP - that their expectations are inserted, PT indicates whether post-training is applied. 
Num.cl corresponds to the number of cases that are classified when using a 0.95 threshold on the probability for making a decision. Density level corresponds to the fraction of weights that are turned on.
All results are medians across $10$ repeated experiments (with min and max included in parentheses). }
  \label{t1}
  \tiny
  \centering
\begin{tabular}{cccc|cccc|cccc}
   \hline
   &&&&\multicolumn{4}{c}{MNIST}&\multicolumn{4}{|c}{FMNIST}\\
\multicolumn{4}{c|}{Method}&\multicolumn{1}{c}{All cl}&\multicolumn{2}{c}{0.95 threshold}&Density
                           &\multicolumn{1}{c}{All cl}&\multicolumn{2}{c}{0.95 threshold}&Density\\
$\bm\gamma$&$\bm\beta$&PT&$R$& Acc& Acc&Num.cl&level
                              & Acc& Acc&Num.cl&level\\
    \hline
    \multicolumn{4}{l|}{\textbf{Full BNN, Gaussian priors}}&&&&&\\\hline
SIM&SIM&N&1&0.958 (0.954,0.960)&-&-&0.056&0.854 (0.850,0.858)&-&-&0.066\\
SIM&SIM&Y&1&0.971 (0.969,0.973)&-&-&0.056&0.868 (0.863,0.872)&-&-&0.066\\
SIM&SIM&N&10&0.967 (0.966,0.971)&0.999 &7064&0.084&0.867 (0.863,0.870)&0.996 &4097&0.083\\
SIM&SIM&Y&10&0.978 (0.976,0.980)&0.999 &8366&0.084&0.880 (0.875,0.882)&0.994 &4933&0.083\\
ALL&EXP&N&1&0.969 (0.967,0.970)&-&-&1.000&0.866 (0.864,0.874)&-&-&1.000\\
ALL&EXP&Y&1&0.979 (0.978,0.980)&-&-&1.000&0.880 (0.877,0.884)&-&-&1.000\\
MED&SIM&N&1&0.961 (0.957,0.964)&-&-&0.051&0.858 (0.854,0.865)&-&-&0.065\\
MED&SIM&Y&1&0.973 (0.971,0.977)&-&-&0.051&0.872 (0.870,0.875)&-&-&0.065\\
MED&SIM&N&10&0.964 (0.962,0.967)&0.998 &7441&0.051&0.863 (0.859,0.869)&0.993 &4347&0.065\\
MED&SIM&Y&10&0.977 (0.976,0.979)&0.999 &8645&0.051&0.878 (0.876,0.881)&0.992 &5223&0.065\\
MED&EXP&N&1&0.965 (0.963,0.968)&-&-&0.051&0.863 (0.859,0.870)&-&-&0.065\\
MED&EXP&Y&1&0.978 (0.976,0.979)&-&-&0.051&0.879 (0.876,0.882)&-&-&0.065\\
\hline
\multicolumn{4}{l|}{\textbf{Dense BNN, Gaussian priors}}&&&&&\\\hline
ALL&SIM&N&1&0.965 (0.965,0.966)&-&-&1.000&0.864 (0.863,0.866)&-&-&1.000\\
ALL&SIM&N&10&0.984 (0.982,0.985)&0.999 &8477&1.000&0.893 (0.890,0.894)&0.997 &5089&1.000\\
ALL&EXP&N&1&0.984 (0.982,0.985)&-&-&1.000&0.886 (0.882,0.888)&-&-&1.000\\
\hline
\multicolumn{4}{l|}{\textbf{Dense BNN, mixture priors}}&&&&&\\\hline
ALL&SIM&N&1&0.965 (0.964,0.967)&-&-&1.000&0.867 (0.866,0.868)&-&-&1.000\\
ALL&SIM&N&10&0.982 (0.981,0.983)&0.999 &8329&1.000&0.893 (0.892,0.897)&0.996 &5151&1.000\\
ALL&EXP&N&1&0.983 (0.981,0.984)&-&-&1.000&0.888 (0.885,0.890)&-&-&1.000\\
\hline
\multicolumn{4}{l|}{\textbf{BNN, Concrete dropout}}&&&&&\\\hline
SIM&SIM&N&1&0.982 (0.894,0.984)&-&-&0.226&0.896 (0.820,0.902)&-&-&0.094\\
SIM&SIM&N&10&0.984 (0.896,0.986)&0.995 &9581&0.820&0.897 (0.823,0.901)&0.942 &8825&0.447\\
SIM&SIM&Y&1&0.982 (0.894,0.984)&-&-&0.226&0.897 (0.820,0.899)&-&-&0.094\\
SIM&SIM&Y&10&0.984 (0.896,0.986)&0.995 &9586&0.820&0.897 (0.823,0.902)&0.943 &8826&0.447\\
ALL&EXP&N&1&0.983 (0.896,0.984)&-&-&1.000&0.896 (0.821,0.901)&-&-&1.000\\
ALL&EXP&Y&1&0.983 (0.894,0.984)&-&-&1.000&0.896 (0.820,0.901)&-&-&1.000\\
\hline
\multicolumn{4}{l|}{\textbf{BNN, horseshoe priors}}&&&&&\\\hline
SIM&SIM&N&1&0.964 (0.962,0.967)&-&-&1.000&0.864 (0.863,0.869)&-&-&1.000\\
SIM&SIM&N&10&0.982 (0.981,0.983)&1.000&0003&1.000&0.887 (0.886,0.889)&1.000 &0181&1.000\\
ALL&EXP&N&1&0.966 (0.963,0.968)&-&-&1.000&0.867 (0.861,0.868)&-&-&1.000\\
PRN&SIM&N&1&0.965 (0.962,0.969)&-&-&0.194&0.865 (0.860,0.868)&-&-&0.302\\
PRN&SIM&N&10&0.982 (0.981,0.983)&1.000 &0002&0.194&0.887 (0.884,0.888)&1.000 &0179&0.302\\
PRN&EXP&N&1&0.965 (0.963,0.968)&-&-&0.194&0.865 (0.862,0.869)&-&-&0.302\\
PRN&SIM&Y&1&0.967 (0.965,0.968)&-&-&0.194&0.867 (0.864,0.871)&-&-&0.302\\
PRN&SIM&Y&10&0.982 (0.981,0.983)&1.000 &0007&0.194&0.888 (0.887,0.890)&1.000 &0147&0.302\\
PRN&EXP&Y&1&0.966 (0.964,0.969)&-&-&0.194&0.868 (0.864,0.869)&-&-&0.302\\
\bottomrule
  \end{tabular}
\end{table}
\begin{table}[h]
 \caption{\small Medians and standard deviations of the average (per layer) marginal inclusion probability (see the text for the definition) for our model for both MNIST and FMNIST data across 10 repeated experiments.}
 \label{t3}
  \tiny
  \centering
  \begin{tabular}{l@{\hspace{0.4cm}}c@{\hspace{0.4cm}}ccc}
   \toprule
&\multicolumn{2}{l}{\textbf{MNIST data}}&\multicolumn{2}{l}{\textbf{FMNIST data}}\\
\hline
Layer& Med.  & SD.& Med.  & SD.\\\hline
$\rho(\gamma^{(1)}|\D)$&0.0520&0.0005&0.0665&0.0004\\
$\rho(\gamma^{(2)}|\D)$&0.0598&0.0003&0.0613&0.0005\\
$\rho(\gamma^{(3)}|\D)$&0.2217&0.0064&0.2013&0.0051\\
\hline
\bottomrule
  \end{tabular}

\end{table}
 \paragraph{MNIST} The results reported in Tables~\ref{t1} and~\ref{t3} show that within the suggested fully Bayesian approach: a) model averaging across different BNNs ($R=10$) gives significantly higher accuracy than the accuracy of a random individual BNN from the model space ($R=1$), additional accuracy improvement can be achieved by using more samples from the joint posterior for model averaging as shown in Figure~\ref{Fig:predsamples}; b) the median probability model and posterior mean based model also perform significantly better than a randomly sampled model, they perform somewhat worse than model averaging, but equally well to model averaging when post-training is performed; c) the majority of the weights of the models have very low marginal inclusion probabilities for the weights at layers 1 and 2, and significantly more weights have higher marginal inclusion probabilities at layer 3, resembling the structure of convolutional neural networks (CNN) where typically one first has a set of sparse convolutional layers, followed by a few fully connected layers, unlike CNNs the structure of sparsification is learned automatically within our approach (see also Section~\ref{ap:sparse} in the web supplement for further details on this issue); d) variations of all of the performance metrics across simulations are low, showing stable behavior across the repeated experiments; 
 e) inference with a doubt option gives almost perfect accuracy, however this comes at a price of rejecting to classify some of the items. 

For other approaches it is also the case that f) both posterior mean based model and using sample averaging improves accuracy compared to a single sample from the parameter space; g) variations of the target parameters are low for the dense BNNs with Gaussian/mixture of Gaussians priors and BNN with horseshoe priors and rather high for the Concrete dropout approach. When it comes to comparing our approach to baselines we notice that h) dense approaches outperform sparse approaches in terms of the accuracy in general; i) Concrete dropout marginally outperforms other sparse approaches in terms of median accuracy, however it exhibits large variance, whilst our full BNN and the compressed BNN with horseshoe priors yield equivalent performance across experiments; j) neither our approach nor baselines managed to reach state of the art results in terms of hard classification accuracy of predictions \citep{palvanov2018comparisons}; k) including a 95\% threshold for making a classification results in a very low number of classified cases for the horseshoe priors (it is extremely underconfident),  the Concrete dropout approach seems to be overconfident when doing inference with the doubt option (resulting in lower accuracy but larger number of decisions), the full BNN, and BNN with Gaussian and mixture of Gaussian priors give less classified cases than the Concrete dropout approach but reach significantly higher accuracy; l) this might mean that the thresholds need to be calibrated towards the specific methods; m) our approach yields the highest sparsity of weights, both for model averaging and for the median probability model.

\paragraph{FMNIST}\label{FMNIST}
The same set of approaches, model specifications and tuning parameters of the algorithms as in the MNIST example were used for this application. The results a)- m) for FMNIST data, based on Tables~\ref{t1},~\ref{t3} and Figure~\ref{Fig:predsamples} are completely consistent with the results from the MNIST experiment, however predictive performance for all of the approaches is poorer on FMNIST. Also whilst full BNN and BNN with horseshoe priors on FMNIST obtain lower sparsity levels than on MNIST, Concrete dropout here improves in this sense compared to the previous example. 

\paragraph{Out-of-sample experiments}

\begin{figure}[t]
\centering
\includegraphics[width=0.49\linewidth]{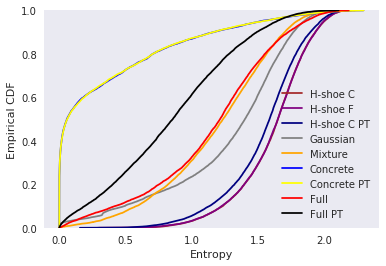}
\includegraphics[width=0.49\linewidth]{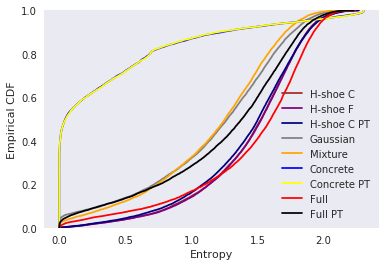}
\centering
\caption{Empirical CDF for the entropy of the marginal posterior predictive distributions trained on MNIST and applied to FMNIST (left) and vice versa (right) for simulation $s=10$. Postfix PT indicates whether post-training is applied, C - if pruning is done, F - if no pruning is done. }\label{Fig:entrop}
\end{figure}

Figure~\ref{Fig:entrop} shows the results on out-of-sample experiments using FMNIST data on a BNN trained by MNIST data and vice versa. Following~\citet{louizos2017multiplicative}, the goal now is to obtain as inconclusive results as possible (reaching ideally a uniform distribution across classes), corresponding to a large entropy. The plot shows the empirical cumulative distribution function  (CDF) of the entropies over the classified samples. 
Concrete drop-out is over confident with a distribution of test classes being far from uniform, the horseshoe prior based approach is the closest to uniform (but it was also closer to uniform for the in-domain predictions), whilst the 2 other baselines are in between, our approach is on par with them for the first case (left graph) and even becomes close to the horseshoe prior based approach for the second case (right graph) showing that it handles out-of-sample uncertainty rather well (just as it does handle the in-domain uncertainty). See also Section~\ref{mnistsup} in the web supplement for further details on this issue.


\section{Discussion}\label{section5}

In this paper we have introduced the concept of Bayesian model (or structural) uncertainty in BNNs and suggested a scalable variational inference technique for fitting the approximation to the joint posterior of models and parameters of these models. Posterior predictive distributions, with both models and parameters marginalized out, can be easily obtained. Furthermore, marginal inclusion probabilities suggested in our approach give proper probabilistic interpretation to Bayesian binary dropout and allow to perform model selection. This comes at the price of having just one additional parameter per weight included. We further provide image classification applications of the suggested technique showing that it both allows to significantly sparsify neural networks without noticeable loss of predictive power and accurately handle the predictive uncertainty.

Currently, fairly simple prior distributions for both models and parameters are used. These prior distributions are assumed independent across the parameters of the neural network, which might not always be reasonable. Alternatively, both parameter and model priors can incorporate joint dependent structures, which can further improve the sparsification of the configurations of neural networks. When it comes to the model priors with local structures and dependencies between the variables (neurons), one can mention the so called dilution priors \citep{george2010dilution}. These priors take care of the similarities between models by means of down-weighting the probabilities of the models with highly correlated variables. 
There are also numerous approaches to incorporate interdependencies between the model parameters via priors in different settings \citep{smith2004bayesian, fahrmeir2001bayesian, dobra2004sparse}. Obviously, in the context of inference in the joint parameter-model settings in BNNs, more research should be done on the choice of priors. Specifically, for image analysis, it might be of interest to develop convolution-inducing priors, whilst for recurrent models one can think of exponentially decaying parameter priors for controlling the short-long memory. 


In this work we restrict ourselves to a subclass of BNNs, defined by inclusion-exclusion of particular weights within a given architecture. In the future it can be of particular interest to extend the approach to the choice of the activation functions as well as the maximal depth and width of each layer of the BNN. A more detailed discussion of these possibilities and ways to proceed is given in \citet{hubin2018thesis}. 
Finally, studies of the accuracy of variational inference within these complex nonlinear models should be performed. Even within linear models \citet{carbonetto2012scalable} have shown that the results can be strongly biased. Various approaches for reducing the bias in variational inference are developed. One can either use more flexible families of variational distributions by for example introducing auxiliary variables \citep{ranganath2016hierarchical, salimans2015markov} or address Jackknife to remove the bias \citep{nowozin2018debiasing}.

{\small
\bibliographystyle{agsm}
\bibliography{lit}
}
 \subsubsection*{Acknowledgments}
The authors would like to acknowledge Sean Murray (Norwegian Computing Center) for the comments on the language of the article and Dr. Pierre Lison (Norwegian Computing Center) for thoughtful discussions of the literature, potential applications and technological tools. We also thank Dr. Petter Mostad, Department of Mathematical Sciences,
Chalmers University of Technology and the University of Gothenburg for useful comments on Proposition 2.1.

\newpage

\appendix

\section*{WEB SUPPLEMENT}
\section{Proofs of propositions}\label{proofs}
\begin{proof}[Proof of \textbf{Proposition 1}]
\vspace*{-0.2cm}
We have 
\begin{align*}
\text{KL}(&q_{\boldsymbol\eta}(\boldsymbol\theta,\boldsymbol\gamma)||p(\boldsymbol\theta,\boldsymbol\gamma|\D)) 
= \sum_{\boldsymbol\gamma \in \Gamma}\int_{\Theta}q_{\boldsymbol\eta}(\boldsymbol\theta,\boldsymbol\gamma)\log\tfrac{q_{\boldsymbol\eta} (\boldsymbol\theta,\boldsymbol\gamma)p(\D)}{p(\D|\boldsymbol\theta,\boldsymbol\gamma)p(\boldsymbol\theta,\boldsymbol\gamma)}\text{d}\boldsymbol\theta\\
=& \log{p(\D)}+\sum_{\boldsymbol\gamma \in \Gamma}\int_{\Theta}q_{\boldsymbol\eta}(\boldsymbol\theta,\boldsymbol\gamma) \log \tfrac{q_{\boldsymbol\eta} (\boldsymbol\theta,\boldsymbol\gamma)}{p(\boldsymbol\theta,\boldsymbol\gamma)}\text{d}\boldsymbol\theta-\sum_{\boldsymbol\gamma \in \Gamma}\int_{\Theta}q_{\boldsymbol\eta}(\boldsymbol\theta,\boldsymbol\gamma)\log{p(\D|\boldsymbol\theta,\boldsymbol\gamma)}\text{d}\boldsymbol\theta \\
=&\log{p(\D)}-\mathcal{L}_{VI}(\boldsymbol\eta).
\end{align*}
from which the result follows.
\end{proof}
\begin{proof}[Proof of \textbf{Proposition 2}]
Consider the continuous relaxation and reparametrization of the model
\begin{align*}
\tilde\gamma_{kj}^{(l)}
=&
\text{sigmoid}({(\text{logit}(\alpha_{kj}^{(l)})-\text{logit}(\nu_{kj}^{(l)}))/\delta}),\quad
\beta_{kj}^{(l)}=\mu_{kj}^{(l)}+\sigma_{kj}^{(l)}\varepsilon_{kj}^{(l)}
\intertext{and similarly an approximative prior distribution given by}
\tilde{\gamma}_{kj}^{(l)}
=&
\text{sigmoid}({(\text{logit}(a)-\text{logit}(\nu_{kj}^{(l)}))/\delta}),\quad
{\beta}_{kj}^{(l)}=\sigma_{\beta}\varepsilon_{kj}^{(l)}
\end{align*}
with $\nu_{kj}^{(l)}\sim\text{Unif}[0,1]$ and $\varepsilon_{kj}^{(l)}\sim N(0,1)$.
For $\delta\rightarrow 0$ this model converges to the original one. Note however that in this model we have a continuous dependence on the $\alpha$'s. Direct application of the pathwise derivative estimation \citep{blundell2015weight, Gal2016Uncertainty, ghosh2017model} can then be performed for any fixed value of $\delta$, similarly to \citet{gal2017concrete}. However we are interested in the limiting case.

Define $E_{q}^{\bm\nu,\bm\varepsilon}$ to be the expectation with respect to $(\bm{\nu},\bm{\varepsilon})$, then for this parametrization the \textit{ELBO} becomes:
\begin{align*}
\mathcal{L}^{*}_{VI}(\bm\eta) 
=&\sum_{i=1}^nE_q^{\bm\nu,\bm\varepsilon}\left\{\log p(\bm{y_i}|\bm{x_i},\tilde{\bm\gamma}(\bm\nu;\bm\alpha),\bm\beta(\bm\varepsilon;\bm\mu,\bm\sigma))\right\}-E_q^{\bm\nu,\bm\varepsilon}\left\{\log\frac{q^{\bm\nu,\bm\varepsilon} (\tilde{\bm\gamma}(\bm\nu;\bm\alpha),\bm\beta(\bm\varepsilon;\bm\mu,\bm\sigma))}
                                        {p_\theta(\tilde{\bm\gamma}(\bm\nu;\bm a),\bm\beta(\bm\varepsilon;\bm 0,\bm\sigma_{\beta}))}\right\}.
\end{align*}
 and its gradient can be written as:
\begin{align*}
\nabla&\mathcal{L}^{*}_{VI}(\bm\eta) 
=\sum_{i=1}^nE_q^{\bm\nu,\bm\varepsilon}\left\{\nabla\log p(\bm{y_i}|\bm{x_i},\tilde{\bm\gamma}(\bm\nu;\bm\alpha),\bm\beta(\bm\varepsilon;\bm\mu,\bm\sigma))\right\}-
E_q^{\bm\nu,\bm\varepsilon}\left\{\nabla\log\frac{q^{\bm\nu,\bm\varepsilon} (\tilde{\bm\gamma}(\bm\nu;\bm\alpha),\bm\beta(\bm\varepsilon;\bm\mu,\bm\sigma))}
                                        {p_\theta(\tilde{\bm\gamma}(\bm\nu;\bm a),\bm\beta(\bm\varepsilon;\bm 0,\bm\sigma_{\beta}))}\right\}.
\end{align*}
To approximate the expectation part and the sum over all observations we use Monte-Carlo sampling. Draw
$(\bm\nu^{(m)},\bm\varepsilon^{(m)})\sim q^{\bm\nu,\bm\varepsilon}(\cdot)$ for $m=1,...,M$ and $S$ as a random subsample of size $N$ from $\{1,...,n\}$. The unbiased estimator of the gradient of \textit{ELBO} then becomes:
\begin{align*}
\widetilde{\nabla}&\mathcal{L}^{*}_{VI}(\bm\eta) 
=\frac{n}{NM}\sum_{i\in S}\sum_{m=1}^M[\nabla\log p(\bm{y_i}|\bm{x_i},\bm\zeta^{(m)})]-
\frac{1}{M}\sum_{m=1}^M\nabla\log\tfrac{q^{\bm\nu,\bm\varepsilon} (\bm\zeta^{(m)})}{p_\theta(\bm\zeta^{(m)})}
\end{align*}
where $\bm\zeta^{(m)}=\bm h(\bm\nu^{(m)},\bm\varepsilon^{(m)};\bm\eta)$. 
When $\delta\rightarrow 0$, estimator $\widetilde{\nabla} {\mathcal{L}}^{*}_{VI}(\boldsymbol\eta)$ converges to $\widetilde{\nabla} {\mathcal{L}}_{VI}(\boldsymbol\eta)$.
\footnote{Note that the gradient of the last term $\text{KL}({q_{\boldsymbol\eta} (\boldsymbol\theta,\boldsymbol\gamma)}||{p(\boldsymbol\theta,\boldsymbol\gamma)})$ in ${\mathcal{L}}_{VI}(\bm\eta)$ actually could have been derived directly, leading to the same results.} 
\end{proof}
\begin{proof}[Proof of \textbf{Remark 2.1}] The identity
\[
\int_{\Theta} q_{\boldsymbol\eta}(\boldsymbol\theta,\boldsymbol\gamma)\nabla  \log q_{\boldsymbol\eta}(\boldsymbol\theta,\boldsymbol\gamma) C_{\boldsymbol\psi}(\boldsymbol{x})\text{d}\boldsymbol\theta = C_{\boldsymbol\psi}(\boldsymbol{x})\nabla\int_{\Theta} q_{\boldsymbol\eta}(\boldsymbol\theta,\boldsymbol\gamma)\text{d}\boldsymbol\theta=0
\]
leads to the result.
\end{proof}
\section{Other inference possibilities}\label{OtherInf}

\subsection*{Model-parameter posterior mean based inference}

Approximate $p(\Delta|\D)$ as
$p(\Delta|\hat E\{\boldsymbol\beta|\D\})$,
where $\hat E\{\beta^{(l)}_{kj}|\D\}= \hat E\{\beta^{(l)}_{kj}|\gamma^{(l)}_{kj}=1,\D\}\hat p(\gamma^{(l)}_{kj}=1|\D) + \hat E\{\beta^{(l)}_{kj}|\gamma^{(l)}_{kj}=0,\D\}\hat p(\gamma^{(l)}_{kj}=0|\D)$, which within our variational Bayes approach simplifies to $\hat E\{\beta^{(l)}_{kj}|\D\} = \alpha^{(l)}_{kj}\mu^{(l)}_{kj}$. Here no sampling (as in Algorithm~\ref{alg:pred}) is needed, but no sparsification is achieved either.






\subsection*{Median probability model based inference combined with sampling}

This approach is based on the notion of a median probability model, which was shown to be optimal in terms of predictions in the context of simple linear models \citep{barbieri2004optimal}. 
This corresponds to applying Algorithm~\ref{alg:pred} with the sampling  of $\gamma_{kj}^{(l)}$ replaced by putting $\gamma_{kj}^{(l)}=\text{I}(\alpha^{(l)}_{kj} > 0.5)$.
Within this approach we significantly sparsify the network and only sample from the distributions of those weights that have marginal inclusion probabilities above 0.5. The rest of the weights are simply replaced with deterministic values of zero.

\subsection*{Median probability model based inference combined with parameter posterior mean}
Approximate $p(\Delta|\D)$ as
$p(\Delta|\hat E\{\boldsymbol\beta|\D\})$,
\text{where} $\hat E\{\beta^{(l)}_{kj}|\D\} = \hat E\{\beta^{(l)}_{kj}|\gamma^{(l)}_{kj}=1,\D\}\text{I}(\hat p(\gamma^{(l)}_{kj}=1|\D)>0.5)$, which within our variational Bayes approach simplifies to $\hat E\{\beta^{(l)}_{kj}|\D\} = \text{I}(\alpha^{(l)}_{kj}>0.5)\mu^{(l)}_{kj}$. Here again no sampling is needed
and we only need to store the variational parameters of  $\widehat{\boldsymbol\eta}$ corresponding to marginal inclusion probabilities above 0.5. Hence we are significantly sparsifying the BNN of interest and reduce the computational cost of inference drastically. 

\subsection*{Median probability model: Post-training}
Once it is decided to make inference based on the median probability model, one might take a number of additional iterations of the training algorithm with respect to the parameters of the models, having the architecture fixed.  This  might often give additional improvements in terms of the quality of inference as well as make the training steps much easier, since the number of parameters is reduced dramatically. This is so, since one does not have to estimate marginal inclusion probabilities $\boldsymbol\alpha$ 
any longer. Moreover, the number of weights $\beta_{jk}^{(l)}$'s corresponding to $\gamma_{jk}^{(l)}=1$ 
to make inference on is significantly reduced due to the sparsity induced by using the median probability model. 

\subsection*{Infeasibility remark: Other model selecting criteria and alternative thresholding}

The median probability model is not always feasible in the sense that one needs at least one connected path across all of the layers with all of the weights linking the neurons having marginal inclusion above 0.5. One way to resolve the issue is to use the most probable model (model with the largest marginal posterior probability) instead of the median probability model. Then conditionally on its configuration one can sample from the distribution of the parameters, select the mean (mode) of the parameters or post-train the distributions of the parameters. Other model selection criteria including DIC, WAIC, and FIC \citep{claeskens2008model} can be used in the same way as the most probable model. Another heuristic way to tackle the issue is to replace conditioning on $\text{I}(\alpha^{(l)}_{kj}>0.5)$ with $\text{I}(\alpha^{(l)}_{kj}>\lambda)$, where $\lambda$ is an arbitrary threshold. The latter will also improve predictive performance in case too conservative priors on the model configurations are used.

\section{Extensions of the applications}\label{applications}

\subsection{Sparsity in the posterior}\label{ap:sparse}
\begin{figure}[t]
\begin{minipage}[t]{1\linewidth}
\begin{minipage}[t]{0.33\linewidth}
\centering
Layer 1

\includegraphics[width=1\linewidth]{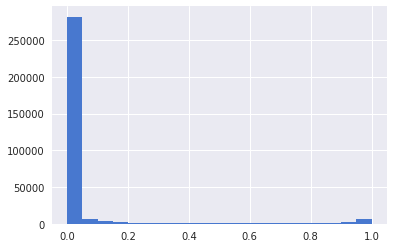}
\includegraphics[width=1\linewidth]{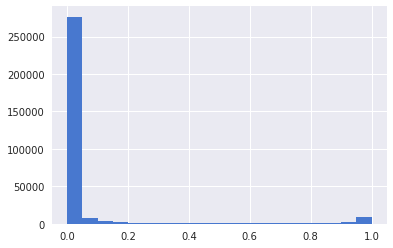}

\end{minipage}
\begin{minipage}[t]{0.33\linewidth}
\centering
Layer 2

\includegraphics[width=1\linewidth]{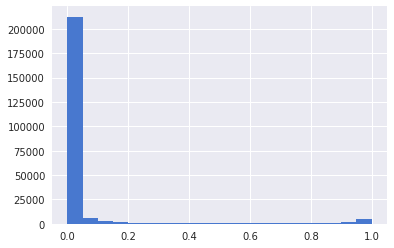}
\includegraphics[width=1\linewidth]{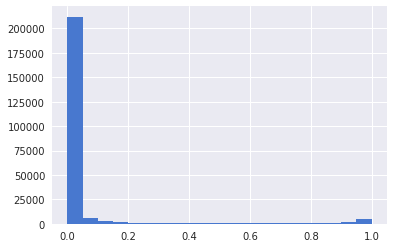}

\end{minipage}
\begin{minipage}[t]{0.32\linewidth}
\centering
Layer 3

\includegraphics[width=1\linewidth]{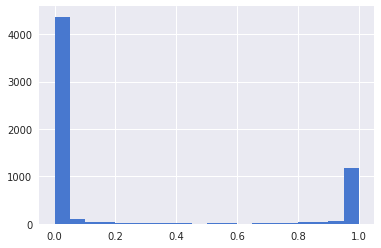}
\includegraphics[width=1\linewidth]{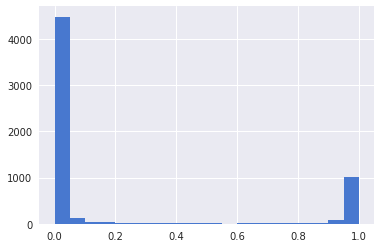}

\end{minipage}
\end{minipage}
\caption{Histograms of the marginal inclusion probabilities of the weights for the three hidden layers from simulation $s=10$ for MNIST (up) and FMNIST (down).}\label{Fig:hist}
\end{figure}

In the top row of Figure~\ref{Fig:hist} we are reporting histograms of the marginal inclusion probabilities for the weights at all of the three hidden layers for MNIST data. Just like in Table~\ref{t3} the histograms show that our approach yields extremely high sparsification levels for layers 1 and 2 and a more moderate sparsification at layer 3.
A similar structure is seen in the bottom row for the FMNIST data.

\begin{figure}[t]
\begin{minipage}[t]{1.0\linewidth}
\centering
Layer 1

\includegraphics[width=0.46\linewidth]{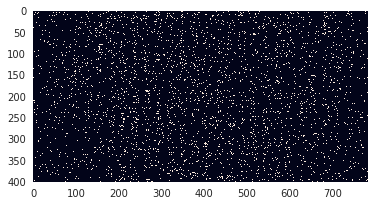}
\includegraphics[width=0.46\linewidth]{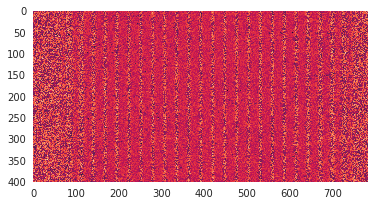}

\end{minipage}

\begin{minipage}[t]{1.0\linewidth}
\centering
Layer 2

\includegraphics[width=0.25\linewidth]{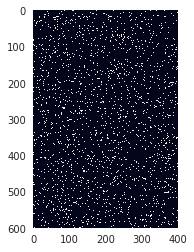}
\includegraphics[width=0.25\linewidth]{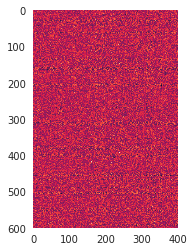}

\end{minipage}

\begin{minipage}[t]{1.0\linewidth}
\centering
Layer 3

\includegraphics[width=0.46\linewidth]{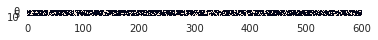}
\includegraphics[width=0.46\linewidth]{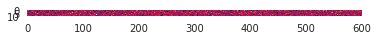}

\end{minipage}
\caption{Random samples from the model space (left) and the weight matrices (right) showing the structure for the three hidden layers from one simulation ($s=10$) for MNIST data.}\label{Fig:strm}
\end{figure}

\begin{figure}[h]

\begin{minipage}[t]{1.0\linewidth}
\centering
Layer 1

\includegraphics[width=0.46\linewidth]{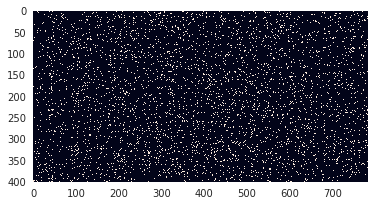}
\includegraphics[width=0.46\linewidth]{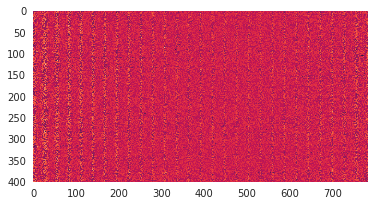}

\end{minipage}

\begin{minipage}[t]{1.0\linewidth}
\centering
Layer 2

\includegraphics[width=0.25\linewidth]{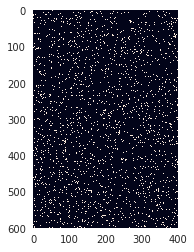}
\includegraphics[width=0.25\linewidth]{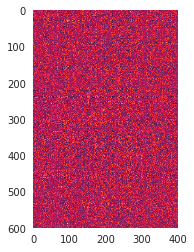}

\end{minipage}

\begin{minipage}[t]{1.0\linewidth}
\centering
Layer 3

\includegraphics[width=0.46\linewidth]{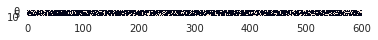}
\includegraphics[width=0.46\linewidth]{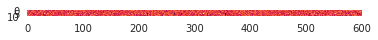}

\end{minipage}
\caption{Random samples from the model space (left) and the weight matrices (right) showing the  structure for the three hidden layers from one simulation ($s=10$) for FMNIST data.}\label{Fig:strf}
\end{figure}

Figures~\ref{Fig:strm} and \ref{Fig:strf} show the sparse structure of typically sampled models from the model space and the corresponding weights. The final sample from the joint posterior is obtained by multiplying the model mask and weights element-wisely. There seems to be some pattern in the structure of the obtained weights for layer 1. This requires additional research and might allow to further compress the BNN. 

\subsection{Miss-classification uncertainties}\label{ap:misclas}

\begin{figure}[p]
\centering
\includegraphics[width=0.49\linewidth]{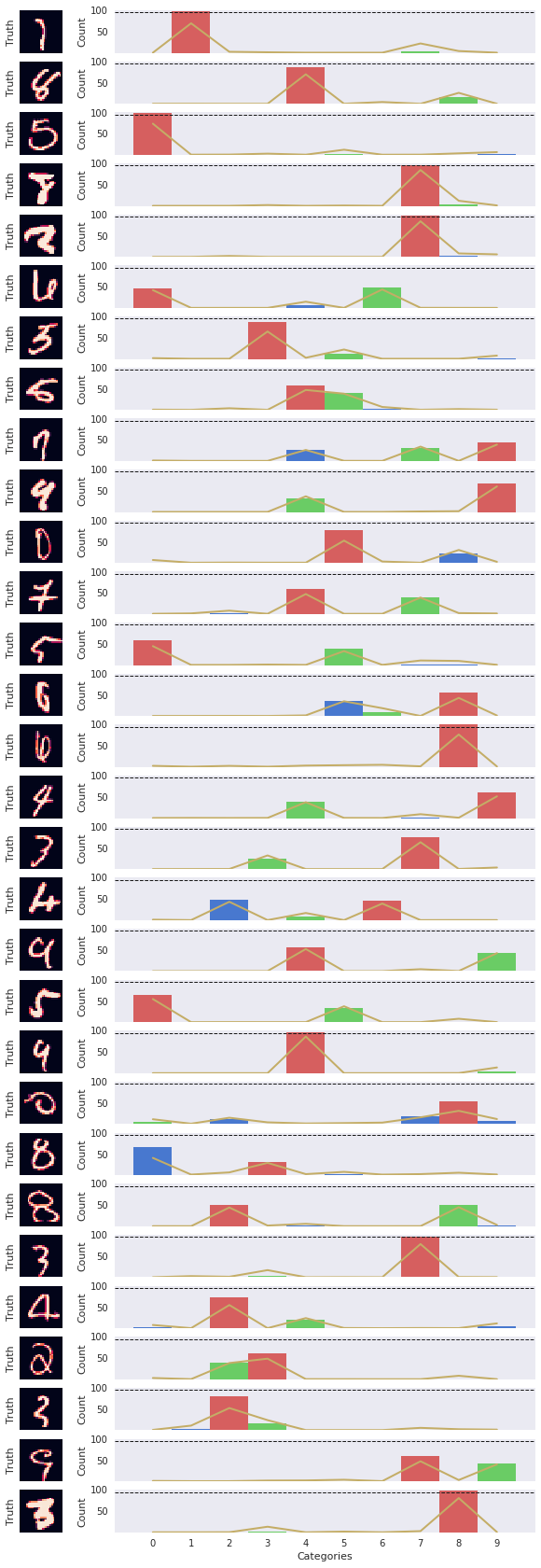}
\includegraphics[width=0.49\linewidth]{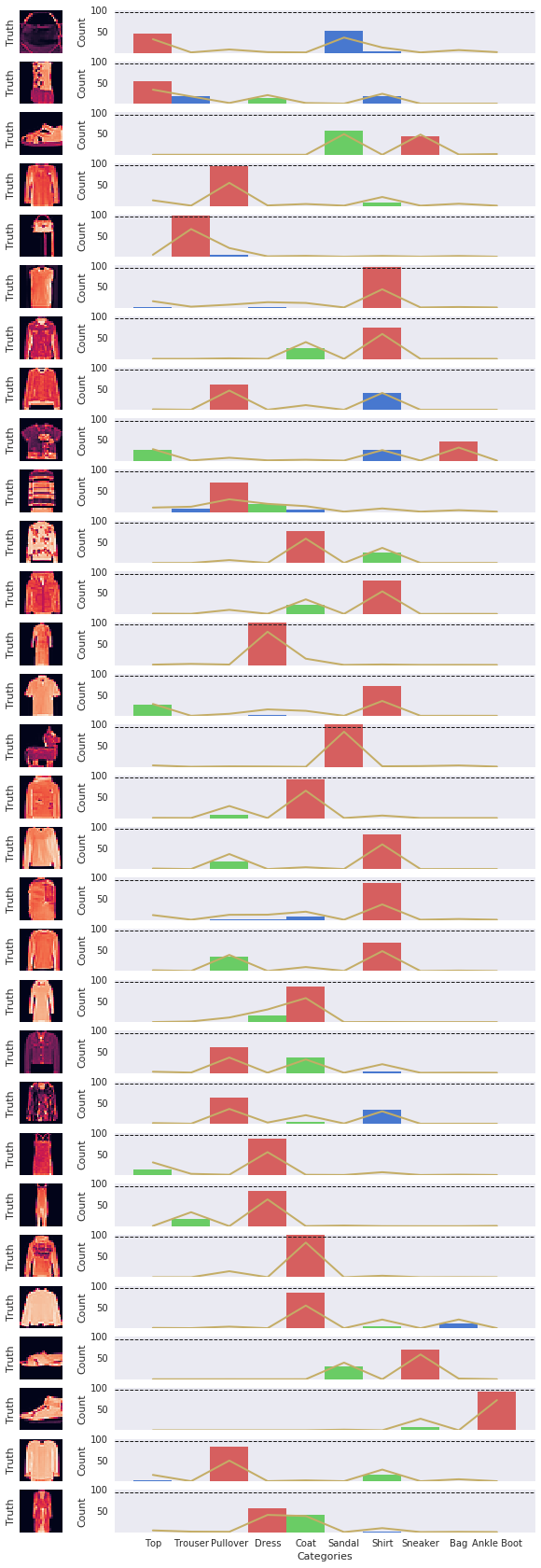}
\caption{Uncertainty based on the samples from the joint posterior (from simulation $s=10$) for 30 wrongly  classified (under model averaging) images for MNIST data (left) and FMNIST data (right). 
 Yellow lines are model averaged posterior class probabilities based on the Full BNN approach (in percent). 
Here green bars are the true classes and red - the incorrectly predicted, blue bars - other samples, dashed black lines indicate the
95\% threshold for making a decision when a doubt possibility is included. The original images are depicted to the left.}\label{Fig:missclass}
\end{figure}

Figure~\ref{Fig:missclass} shows the miss-classification uncertainties associated with posterior predictive sampling. One can clearly see that for almost all of the cases, when the BNN makes a miss-classification, class certainty of the predictions is very low, indicating that the network is unsure. Moreover, even in these cases the truth is typically within the 95\% credible interval of the predictions, which following \citet{posch2019variational} can be read from whether less than 95 out of 100 samples belong to a wrong class and at least 6 out of 100 samples belong to the right one. Also notice that in many of the cases of miss-classification illustrated here, even a human would have serious doubts in making a decision. 

\subsection{In-out-of-domain results}\label{mnistsup}
\begin{figure}[t]
\centering
\includegraphics[width=0.49\linewidth]{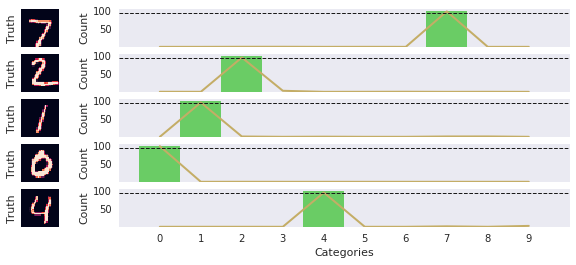}
\includegraphics[width=0.49\linewidth]{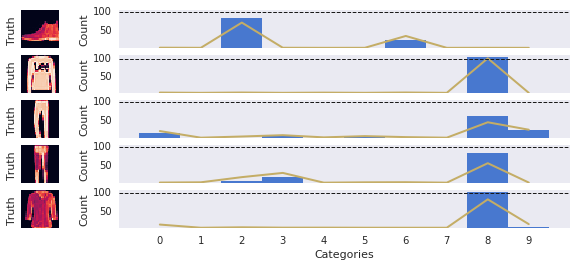}
\caption{Uncertainty related to the in-domain test data (MNIST, left) and out-of-domain test data (FMNIST, right) based on 100 samples from the posterior predictive distribution.  Yellow lines are model averaged posterior class probabilities (in percent).
Green bars mark the correct classes, blue bars for other samples (with heights corresponding to an alternative estimate of class-probabilities using hard classification within each of the replicates in Algorithm~\ref{alg:pred}. The dashed black lines give the 95\% threshold for making decisions with doubt possibilities).
The original images are depicted to the left.}\label{Fig:uncer}
\end{figure}

\begin{figure}[H]
\centering
\includegraphics[width=0.49\linewidth]{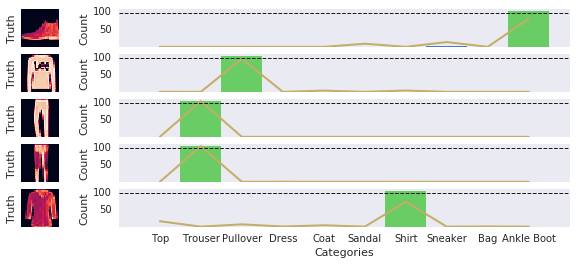}
\includegraphics[width=0.49\linewidth]{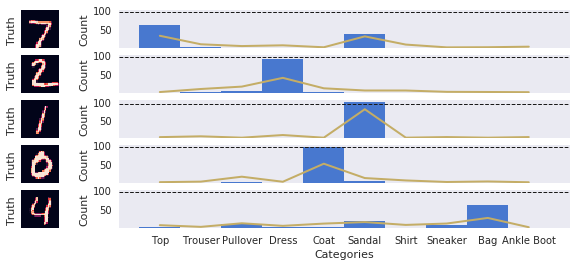}
\caption{Uncertainty related to the in-domain test data (FMNIST, left) and out-of-domain test data (MNIST, right) based on 100 samples from the posterior predictive distribution. See Figure~\ref{Fig:uncer} for additional details.}\label{Fig:funcer}
\end{figure}

Following the example of measuring the in and out-of-domain uncertainty suggested in \citet{blogwu} we will test the ability of the approach to give confidence in its predictions by means of trying to classify a sample from FMNIST images with samples from the posterior predictive distribution based on the joint posterior of models and parameters trained on MNIST data set and compare this to the results for a sample of images from the test set of MNIST data. The results are reported for the joint posterior (of models and parameters) obtained in experiment run $s = 10$. As can be seen in Figure~\ref{Fig:uncer}, the samples from BNN give highly confident predictions for the MNIST data set with almost no variance in the samples from the posterior predictive distribution. At the same time, the out-of-domain uncertainty, related to the samples from the posterior predictive distribution based on FMNIST data, is typically high (with some exceptions) showing low confidence of the samples from the posterior predictive distribution in this case. The reversed example of inference on FMNIST and uncertainty related to MNIST data, illustrated in Figure~\ref{Fig:funcer}, leads to exactly the same conclusions. 
\newpage


\end{document}